\documentclass[11pt]{article}

\usepackage{acl}
\usepackage{times}
\usepackage{latexsym}
\usepackage{microtype}
\usepackage{graphicx}
\usepackage{amsmath}
\usepackage{booktabs}
\usepackage{array}
\usepackage{tabularx}
\usepackage{enumitem}
\usepackage[T1]{fontenc}
\usepackage[utf8]{inputenc}
\usepackage{multirow}
\usepackage[table]{xcolor}
\usepackage{dblfloatfix}
\usepackage[skins,breakable]{tcolorbox}
\definecolor{groupbg}{HTML}{E0E0E0}
\definecolor{agentbg}{HTML}{BBDEFB}
\definecolor{promptbg}{HTML}{F8FAFC}
\definecolor{promptframe}{HTML}{8A8A8A}

\newtcolorbox{promptbox}[1]{
  enhanced jigsaw,
  breakable,
  colback=promptbg,
  colframe=promptframe,
  boxrule=0.35pt,
  arc=1mm,
  left=4pt,
  right=4pt,
  top=3pt,
  bottom=3pt,
  before skip=0.45em,
  after skip=0.6em,
  title={#1},
  fonttitle=\bfseries\footnotesize,
  coltitle=black,
  colbacktitle=groupbg,
  attach boxed title to top left={xshift=3pt,yshift=-1.5mm},
  boxed title style={boxrule=0pt,arc=1mm,left=3pt,right=3pt,top=1pt,bottom=1pt},
  fontupper=\footnotesize\raggedright
}

\newtcolorbox{methodbox}[1]{
  enhanced jigsaw,
  breakable,
  colback=agentbg!18,
  colframe=agentbg!70!black,
  boxrule=0.35pt,
  arc=1mm,
  left=5pt,
  right=5pt,
  top=5pt,
  bottom=5pt,
  before skip=0.6em,
  after skip=0.8em,
  title={#1},
  fonttitle=\bfseries\footnotesize,
  coltitle=black,
  colbacktitle=agentbg,
  attach boxed title to top left={xshift=3pt,yshift=-1.5mm},
  boxed title style={boxrule=0pt,arc=1mm,left=3pt,right=3pt,top=1pt,bottom=1pt},
  fontupper=\footnotesize\raggedright
}

\title{When Summaries Distort Decisions: Information Fidelity in LLM-Compressed Financial Analysis}

\author{
\bfseries Hoyoung Lee\textsuperscript{1, 2} \quad Suhwan Park\textsuperscript{1} \quad Seunghan Lee\textsuperscript{3} \quad Jun Seo\textsuperscript{3} \quad Jaehoon Lee\textsuperscript{3} \\
\bfseries Sungdong Yoo\textsuperscript{3} \quad Minjae Kim\textsuperscript{3} \quad CheolWon Na\textsuperscript{4} \quad Zhangyang Wang\textsuperscript{5} \quad V.Zach Golkhou\textsuperscript{6} \\
\bfseries Minkyu Kim\textsuperscript{7} \quad Sotirios Sabanis\textsuperscript{8, 9, 10} \quad Alejandro Lopez-Lira\textsuperscript{11} \quad Dhagash Mehta\textsuperscript{12} \\
\bfseries Soonyoung Lee\textsuperscript{3} \quad Chanyeol Choi\textsuperscript{2} \quad Wonbin Ahn\textsuperscript{3,\textdagger} \quad Yongjae Lee\textsuperscript{1, 2,\textdagger} \\[0.6em]
{\normalfont\normalsize \textsuperscript{1}UNIST \quad \textsuperscript{2}LinqAlpha \quad \textsuperscript{3}LG AI Research \quad \textsuperscript{4}Sungkyunkwan University} \\
{\normalfont\normalsize \textsuperscript{5}University of Texas at Austin \quad \textsuperscript{6}J.P. Morgan Chase \quad \textsuperscript{7}State Street Investment Management} \\
{\normalfont\normalsize \textsuperscript{8}University of Edinburgh \quad \textsuperscript{9}National Technical University of Athens} \\
{\normalfont\normalsize \textsuperscript{10}Archimedes/Athena Research Centre \quad \textsuperscript{11}University of Florida \quad \textsuperscript{12}BlackRock} \\[0.35em]
{\normalfont\normalsize \textsuperscript{\textdagger}Correspondence: wonbin.ahn@lgresearch.ai, yongjaelee@unist.ac.kr}
}

\newcommand{\modelname}[1]{\textsc{#1}}
\AtBeginDocument{%
  \setlength{\abovedisplayskip}{6pt plus 2pt minus 2pt}%
  \setlength{\belowdisplayskip}{6pt plus 2pt minus 2pt}%
  \setlength{\abovedisplayshortskip}{6pt plus 2pt minus 2pt}%
  \setlength{\belowdisplayshortskip}{6pt plus 2pt minus 2pt}%
}

\begin{document}
\maketitle

\begin{abstract}
Financial decision-makers face more information than they can directly inspect, making context compression necessary. Yet when large language models (LLMs) compress financial source material, they can alter the investment judgment supported by the original source. We frame this problem as information fidelity: compression loses fidelity when it changes the decision induced by the source. In agentic systems, such losses may recur across intermediate steps and amplify throughout the decision process. Across financial filings and earnings-call transcripts, we find that LLM-based compression can produce fluent and factually plausible compressed contexts that nevertheless alter downstream decisions. We analyze two diagnostic patterns associated with fidelity loss: decontextualization, where salient evidence is retained but separated from the caveats and contextual qualifiers needed for correct interpretation, and model dependency, where different compressors expose different views of the same source.
We then propose Agentic Context Compression, which generates multiple candidate compressions and audits their disagreements against the original source. Our results suggest that financial compression should be evaluated not only by efficiency or factuality, but also by its ability to preserve decision-relevant context.
\end{abstract}

\section{Introduction}

\begin{figure}[t]
    \centering
    \includegraphics[width=0.86\linewidth]{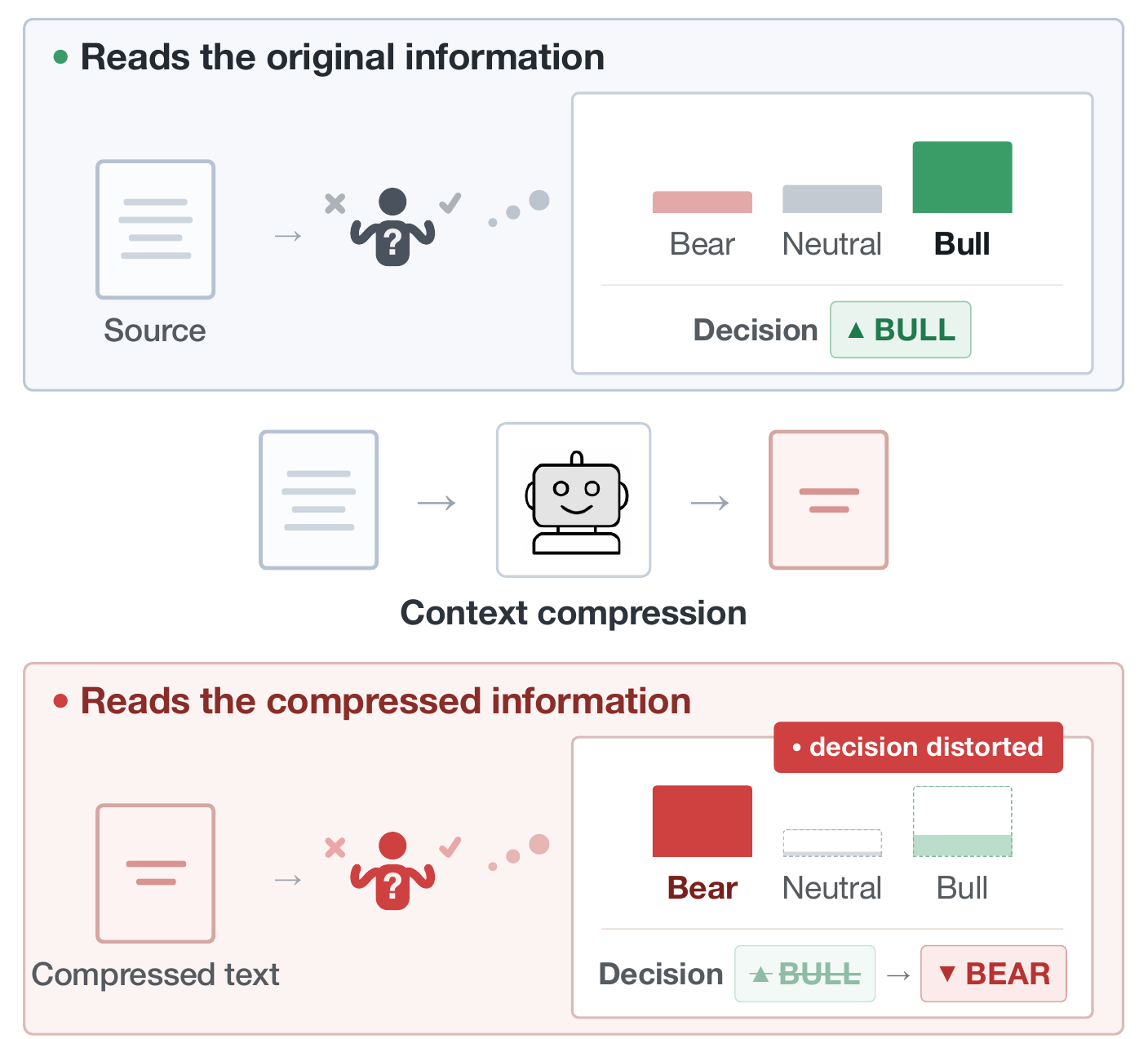}
    \caption{
    \textbf{Compression-induced decision flip.} Context compression can cause a decision-maker to reach a different decision from the one supported by the original source. For example, information that supports a \textit{bullish} decision in the original source may lead to a \textit{bearish} decision after compression.
  }
    \label{fig:motivation}
\end{figure}

Large language models (LLMs) are increasingly used to compress long documents into shorter contexts before those contexts are passed to another model or a human decision-maker. In high-stakes domains such as finance, the resulting short context is useful only if it preserves the source evidence needed to support the decision.

Long-context compression is also inherently open-ended. A financial source text has no single correct summary because different summaries can select different evidence while remaining fluent and plausible. This paper studies a simple but consequential failure mode: an LLM-compressed representation can lead a decision-maker to make a financial judgment that differs from the judgment supported by the full source text, as shown in Figure~\ref{fig:motivation}. The issue is not limited to hallucination or factual inconsistency. It also depends on which evidence remains visible, how it is contextualized, and how competing signals are balanced.

Existing compression evaluations often measure reconstruction, factual consistency, or surface-level information preservation \citep{lajewska2025preservation}, but they do not directly test whether compression preserves the decision supported by the original source. We frame this problem as information fidelity, where a compressed representation has high information fidelity if it preserves the decision implied by the original source. A summary may retain headline facts but omit the caveats or offsets needed to interpret them, remaining factually correct while inducing a different decision.

We examine two structured patterns that help characterize fidelity loss. Decontextualization builds on work showing that isolated claims need context to remain meaningful \citep{choi2021decontext,gunjal2024molecular}. Compression may preserve headline claims while stripping away decision-relevant caveats, comparisons, or expectation framing. When such context is removed, a factually correct summary can still carry a different decision implication.

Model dependency is motivated by evidence that investment analyses vary across LLMs \citep{lee2025yourai}. In our setting, the relevant variation is not only how models analyze the same evidence, but also which evidence each compressor preserves from the same source. These patterns suggest that decision change is not merely random degradation. Instead, it can reflect systematic differences in the evidence made visible after compression.

These observations motivate a source-grounded compression approach that preserves contextual qualifiers and compares multiple compressed views of the same source. We instantiate this approach as agentic context compression, in which an agent generates competing compressed contexts and audits their disagreements against the original source.

We evaluate information fidelity and compare several compression methods on quarterly filings and earnings-call transcripts. Our main contributions are as follows:
\begin{itemize}[leftmargin=*]
\item We introduce \emph{information fidelity}, a decision-centered criterion for evaluating LLM context compression against the original source.

\item Using real-world quarterly filings and earnings-call transcripts, we show that LLM-based compression can alter decisions, with decontextualization and model dependency emerging as recurring diagnostic patterns.

\item We propose agentic context compression, a source-grounded procedure that audits multiple candidate compressed contexts, and show that it reduces compression-induced decision flips.

\end{itemize}

\section{Related Work}

\subsection{Context Compression}

Context compression has become a central problem in LLM-based systems. Prior work shows that LLM-mediated representations can degrade source information during agentic context updating \citep{zhang2026ace} and delegated document workflows \citep{laban2026corrupt}. Compressed contexts may preserve the general topic while dropping essential details, motivating grounding and information-preservation metrics beyond efficiency \citep{lajewska2025preservation}. A related line asks what compression should retain: LLM salience aligns only weakly with human judgment \citep{trienes2025salience}, and conformal summarization formalizes coverage of critical content in high-stakes settings \citep{kuwahara2025conformal}. Even recent generative prompt compression that targets coherence and key information \citep{zhang2025scope} is evaluated mainly by task quality or summary metrics, not by preservation of the source-implied decision.

\subsection{Summarization Distortion}

Repeated generation through LLMs can cumulatively distort source information \citep{mohamed2025broken}. Summarization can likewise distort a source by selectively reweighting content and failing to preserve perspectives or polarity balance. FairSumm measures gaps between source and summary perspectives \citep{zhang2024fairsumm}; opinion summarization can amplify majority polarity and weaken minority polarity \citep{lei2024poca}; and LLMs may implicitly resolve disagreement during generation, suppressing minority viewpoints \citep{aghaebe2026faithful}. LLM-generated news summaries also introduce framing through selective emphasis and omission \citep{pastorino2025framing}.

LLM-generated summaries can also change readers' decisions while remaining fluent and plausible. \citet{alessa2025cognitive} find that LLM-generated content can induce framing and primacy biases and shift human purchase decisions, while \citet{peters2025generalization} find that LLM summaries of scientific studies often omit scope-limiting details and overgeneralize conclusions.

These studies show that summarization is not neutral compression but a process of selecting and aggregating perspectives \citep{mayilvaghanan2025blindspot}, where selective omission is itself a form of bias. We extend this insight to financial texts by asking which decision-relevant evidence survives.

\subsection{LLMs in Financial Decision-Making}

LLMs are increasingly embedded in financial workflows, from semantic stock representations for thematic investing \citep{lee2025theme} and text-paired time-series construction \citep{lee2026fintexts} to screening of trading signals \citep{kim2026leadlag} and alpha extraction from corporate disclosures \citep{choi2025alpha}. At the same time, research on financial LLMs shows that the investment judgments they produce can be systematically biased by company names, asset classes, prior knowledge, or evaluation design. \citet{glasserman2023lookahead} show that company names and general knowledge can interfere with financial sentiment prediction. \citet{nakagawa2024company} quantify company-specific bias in financial sentiment analysis. \citet{lee2025yourai} analyze model-specific investment preferences and confirmation bias in LLM investment analysis. \citet{kong2026bias} place these issues in a broader evaluation framework and argue that financial LLM evaluation must account for bias.

\section{Information Fidelity}

\subsection{Problem Formulation}

\paragraph{Task.}
We define information fidelity as source-relative decision preservation: a compression has high fidelity when the decision-model belief induced by the compressed text remains close to the belief induced by the original source. Given a source text $s$, a compression model $\mathcal{M}$ produces a compressed text $c$ whose length $\lvert c\rvert$, measured in bullets, must respect a fixed budget $B$:
\begin{equation}
c = \mathcal{M}(s)
\quad\text{subject to}\quad
\lvert c\rvert \le B .
\end{equation}
We hold $B$ fixed across methods so that fidelity differences reflect what each method retains rather than how much it writes. The realized token ratio $\tau=\mathrm{tok}(c)/\mathrm{tok}(s)$ is reported descriptively.

A decision model $\mathcal{E}$ maps any text to a belief distribution over the label set $V=\{\text{bear}, \text{neutral}, \text{bull}\}$, that is, $\mathcal{E}(\cdot)\in\Delta(V)$, the probability simplex over $V$. For each fixed text, we average $R=3$ independent decision runs, yielding a source belief $p_s$ and a compressed belief $p_c$,
\begin{equation}
\label{eq:belief}
p_s = \frac{1}{R}\sum_{r=1}^{R} \mathcal{E}^{(r)}(s),
\qquad
p_c = \frac{1}{R}\sum_{r=1}^{R} \mathcal{E}^{(r)}(c),
\end{equation}
where $c=\mathcal{M}(s)$ is fixed and $\mathcal{E}^{(r)}$ is the $r$-th decision-model run. Their top decisions are the most likely labels

\begin{equation}
\hat{v}_x = \operatorname*{arg\,max}_{v\in V} p_x(v),
\qquad x \in \{s, c\}.
\end{equation}

We quantify its loss with two metrics. The Decision Flip rate is the fraction of source documents whose top decision changes after compression,

\begin{equation}
\mathrm{Flip}
= \frac{1}{N}\sum_{i=1}^{N}
\mathbf{1}\!\left[\, \hat{v}_c^{(i)} \neq \hat{v}_s^{(i)} \,\right],
\end{equation}

where $i$ indexes the $N$ source documents, while Total Variation Distance (TVD), $d_{\mathrm{TV}}(p_c,p_s)$, measures how far the belief distribution moves even when the top decision is unchanged,

\begin{equation}
d_{\mathrm{TV}}(p_c,p_s)
= \frac{1}{2}\sum_{v\in V}
\bigl\lvert p_c(v)-p_s(v) \bigr\rvert .
\end{equation}

The flip rate captures outright decision changes, while TVD captures smaller belief shifts that may not cross the decision boundary. We report TVD as its mean over source texts for each method.

\begin{figure*}[t]
    \centering
    \includegraphics[width=\linewidth]{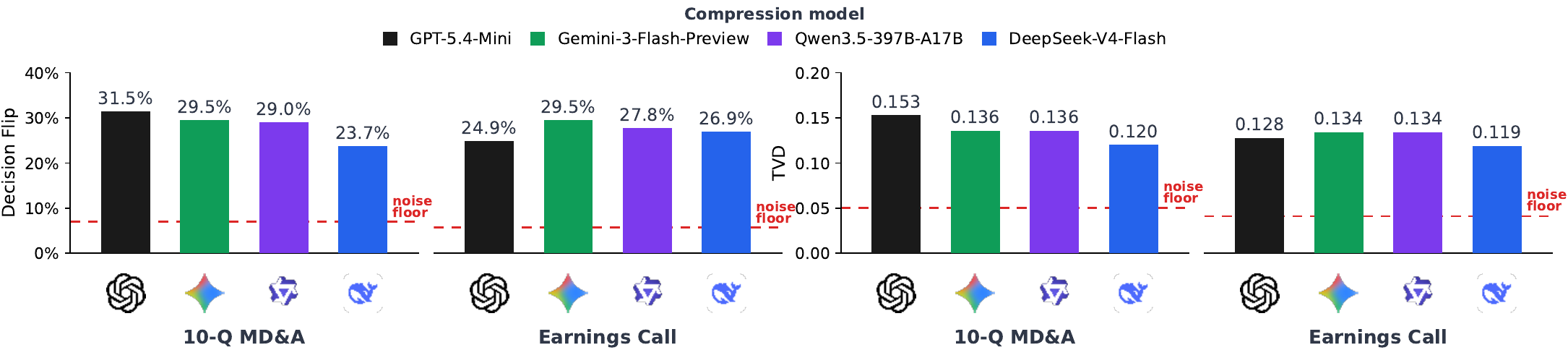}
\caption{Decision change under one-shot compression. Bars show Decision Flip and source-relative TVD for four compressors across MD\&A and earnings calls, averaged across the two decision models. Dashed lines show the decision-model randomness floor: the same decision model reads the original source again, with no compression.}
\label{fig:flow1}
\end{figure*}

\subsection{LLM Context Compression}
\label{subsec:llm-context-compression}

We realize the compression operator $\mathcal{M}$ with calls to a language model. Following \citet{kang2025acon}, we write $\mathrm{LLM}(x;\theta,\mathcal{P})$ for a model with pretrained weights $\theta$ that maps an input $x$ to an output under a natural-language prompt $\mathcal{P}$.

\paragraph{Naive prompt.}
The naive baseline is a single pass $c=\mathrm{LLM}(s;\theta,\mathcal{P}_{\text{comp}})$: the model receives the full source and a fixed bullet budget, then selects the facts it judges most material. The full prompt is shown in Appendix~\ref{app:prompts}.

\paragraph{Contextualization.}
\label{subsec:contextualization}

Contextualization uses the same fixed budget but changes what counts as worth keeping. Instead of choosing facts that look important on their own, it asks the compressor to preserve the details that make each material point interpretable: the relevant comparison, caveat, offsetting signal, or qualifier. Under the same model, source, and budget, it uses $\mathcal{P}_{\text{ctx}}$ to generate candidate contexts $\tilde{c}=\mathrm{LLM}(s;\theta,\mathcal{P}_{\text{ctx}})$. The mechanism is unchanged; only the selection criterion shifts from isolated salience to interpretable evidence.

\subsection{Agentic Context Compression}

We further consider Agentic Context Compression (ACC), which realizes the same operator $\mathcal{M}$ as a two-stage agentic compression process. First, the agent calls LLM candidate-generation tools to produce contextualized candidate contexts $c_k=\mathrm{LLM}(s;\theta_k,\mathcal{P}_{\text{ctx}})$ from the same source $s$, where $k$ indexes a distinct compression model. Second, the agent audits disagreements among the set $\{c_k\}$ against the source. At step $t$, it emits an action $a_t=\mathrm{LLM}(o_t,\mathbf{h}_{t-1};\theta,\mathcal{P}_{\text{agent}})$ from the latest tool observation $o_t$ and history $\mathbf{h}_{t-1}=(o_1,a_1,\dots,o_{t-1},a_{t-1})$. For a claim or omission that differs across candidate contexts $c_k$, the action can grep the source for the relevant terms and then read only a short source snippet around the matching span, following the direct-corpus-interaction view of agentic search \citep{li2026dci}. The agent uses these snippets to test whether a candidate context $c_k$ overclaims the source, for example by making the evidence more bullish, bearish, or certain than the source supports, then scores each $c_k$ by overclaim risk and source-locus coverage, and commits to the intact compressed context that the source best supports (Appendix~\ref{app:agent}). In preliminary experiments, allowing the agent to edit candidate text tended to break the preserved context and degrade decision fidelity, so we restrict the agent to selecting among intact candidates rather than revising them. This connects ACC to multi-agent factuality verification \citep{ning2025madfact} and financial atomic-claim grounding \citep{guo2026finground}, but targets decision-changing omissions and overclaims rather than factuality alone.

\section{Experiments}

\subsection{Experimental Setup}

\paragraph{Datasets.}
Our main panel covers two source types from S\&P 100 firms over fiscal 2025 Q1--Q3: 10-Q MD\&A sections ($N=300$), collected from structured SEC filings following EDGAR-Crawler \citep{loukas2025edgarcrawler}, and quarterly earnings call transcripts ($N=297$).

\begin{table}[htbp]
\centering
\small
\renewcommand{\arraystretch}{1.12}
\begin{tabular}{lcc}
\toprule[1.2pt]
 & 10-Q MD\&A & Earnings Call \\
\midrule
Self-agreement & 90.2${\pm}$9.3\% & 90.5${\pm}$7.8\% \\
Fleiss' $\kappa$ & 0.839${\pm}$0.135 & 0.819${\pm}$0.140 \\
\bottomrule[1.2pt]
\end{tabular}
\caption{Reliability of the decision models. We report pairwise top-decision agreement and Fleiss' $\kappa$ across three independent reads of the same raw source, as the mean${\pm}$s.d.\ across the two decision models.}
\label{tab:decision-model-reliability}
\end{table}

\paragraph{Decision models.}
We use two decision models, \modelname{Gemini-3.1-Flash-Lite} and \modelname{Claude-Haiku-4.5}, under the same decision prompt: given a source $s$ or compressed text $c$, each produces the belief distribution used to compute $p_s$ or $p_c$.

\paragraph{Baselines.}
We compare against compression baselines under the same budget. \textbf{Chunking} partitions $s$ into fixed-size chunks, applies the same compression to each chunk, and concatenates the outputs. For \textbf{Token Pruning}, we use LLMLingua \citep{jiang2023llmlingua}, which drops low-information tokens, and LongLLMLingua \citep{jiang2024longllmlingua}, its long-context extension; both use \modelname{Finance-Llama3-8B} as the token-scoring model. \textbf{Integrator} obtains compressed texts from multiple models and combines them into one final context.

\begin{figure*}[t]
    \centering
    \includegraphics[width=\linewidth]{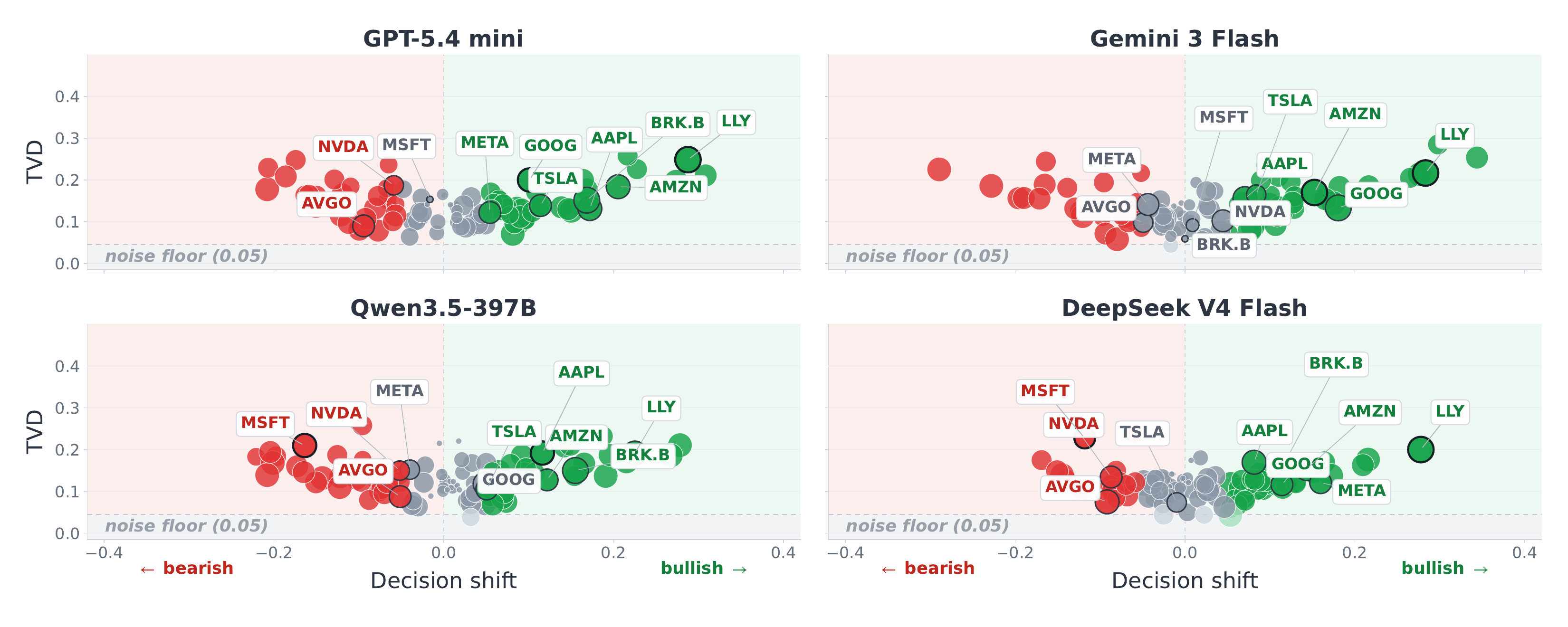}
\caption{Model-specific decision movement under one-shot compression. Each panel shows one compressor; points are source-relative movements per company, pooled over MD\&A and earnings calls and averaged across the two decision models. The horizontal axis shows signed decision shift from bearish to bullish, the vertical axis shows TVD magnitude, and the dashed line marks the noise floor.}
    \label{fig:model-movement}
\end{figure*}

\subsection{Fidelity Loss Under Compression}

Compression changes the decision model's output beyond rereading noise. In Figure~\ref{fig:flow1}, the dashed line is the decision-model randomness floor: the same model rereads the original source, with no compression. Averaged across the two decision models, rereading still flips 7.0\% of MD\&A documents and 5.7\% of earnings calls, so this floor captures rereading randomness alone.

All four one-shot compressors exceed this floor by more than a factor of two on both source types. The excess is the compression-induced component: summaries remove or reweight decision-relevant evidence. The parallel rise in TVD shows that compression also shifts the decision model's belief distribution even without a top-label flip.

Table~\ref{tab:decision-model-reliability} addresses the possibility that these flips come from unstable decision models. Across three independent reads of the same raw source, both models reproduce their top decisions at high rates, with Fleiss' $\kappa$ in the substantial-agreement range or above on both disclosure types. This makes it unlikely that the gap in Figure~\ref{fig:flow1} is only a judge artifact, and supports interpreting the no-compression line as a rereading-noise floor. Per-model results are reported in Appendix~\ref{app:per-evaluator}, and the main conclusions persist when the label space is refined to five return bands (Appendix~\ref{app:five-class}).

\subsection{Model Dependency}

All compressors lose fidelity in aggregate, but flip rate reduces movement to a binary outcome (Figure~\ref{fig:flow1}). We call this model dependency when the size or direction of belief shifts varies with the compressor. To examine this pattern, we plot source-relative movement by ticker for each compressor.

Figure~\ref{fig:model-movement} separates these shifts by compressor. The horizontal axis shows direction, from bearish to bullish, and the vertical axis shows magnitude as source-relative TVD. If compression were model-invariant, the panels would look similar. Instead, compressors occupy different regions: some move tickers toward bearish decisions, others toward bullish decisions, and high-TVD points differ across panels. This pattern suggests that compression-induced decision change is not merely information loss; single-model summaries can carry a model-specific tilt.

If part of the tilt is compressor-specific, aggregating models may reduce it. The Integrator provides a simple check: it differs from single-model compression only by merging a second model's summary, and Table~\ref{tab:main-compression-results} shows this alone lowers the MD\&A flip rate from 31.5\% to 25.6\%. ACC reduces it further by drawing on both models and checking their disagreements against the source.

\begin{table*}[!t]
\centering
\small
\renewcommand{\arraystretch}{1.12}
\setlength{\tabcolsep}{3pt}
\begin{tabularx}{\textwidth}{l
  >{\hsize=1.3\hsize\centering\arraybackslash}X
  >{\hsize=0.9\hsize\centering\arraybackslash}X
  >{\hsize=1.0\hsize\centering\arraybackslash}X
  >{\hsize=0.8\hsize\centering\arraybackslash}X
  >{\hsize=1.3\hsize\centering\arraybackslash}X
  >{\hsize=0.9\hsize\centering\arraybackslash}X
  >{\hsize=1.0\hsize\centering\arraybackslash}X
  >{\hsize=0.8\hsize\centering\arraybackslash}X}
\toprule[1.2pt]
& \multicolumn{4}{c}{10-Q MD\&A}
& \multicolumn{4}{c}{Earnings Call} \\
\cmidrule(lr){2-5} \cmidrule(lr){6-9}
Method
& Flip $\downarrow$
& TVD $\downarrow$
& Cost $\downarrow$
& $\tau$
& Flip $\downarrow$
& TVD $\downarrow$
& Cost $\downarrow$
& $\tau$ \\
\midrule
\textit{No compression}
& 7.0${\pm}$5.6\% & 0.050 & - & 100\%
& 5.7${\pm}$3.3\% & 0.041 & - & 100\% \\
\midrule
\rowcolor{groupbg}
\multicolumn{9}{l}{LLM (\modelname{GPT-5.4-Mini})} \\
\midrule
\quad Naive Prompt
& 31.5${\pm}$2.2\% & 0.153 & \$6.69 & 4.1\%
& 24.9${\pm}$1.4\% & 0.128 & \$3.83 & 4.6\% \\
\quad Chunking
& 30.1${\pm}$9.8\% & 0.133 & \textbf{\$6.03} & 4.0\%
& 29.8${\pm}$2.6\% & 0.138 & \$4.26 & 4.0\% \\
\quad Contextualization
& 21.2${\pm}$0.2\% & \textbf{0.096} & \$8.06 & 4.2\%
& 22.1${\pm}$0.2\% & 0.111 & \textbf{\$3.82} & 4.2\% \\
\midrule
\rowcolor{groupbg}
\multicolumn{9}{l}{Token Pruning (\modelname{Finance-Llama3-8B})} \\
\midrule
\quad LLMLingua
& 49.8${\pm}$4.5\% & 0.218 & - & 4.0\%
& 59.7${\pm}$12.6\% & 0.276 & - & 4.0\% \\
\quad LongLLMLingua
& 43.0${\pm}$9.9\% & 0.200 & - & 4.0\%
& 40.8${\pm}$2.4\% & 0.191 & - & 4.0\% \\
\midrule
\rowcolor{groupbg}
\multicolumn{9}{l}{Multi-LLM (\modelname{GPT-5.4-Mini} + \modelname{DeepSeek-V4-Flash})} \\
\midrule
\quad Integrator
& 25.6${\pm}$1.3\% & 0.121 & \$9.76 & 4.3\%
& 21.7${\pm}$1.7\% & 0.114 & \$6.52 & 4.7\% \\
\rowcolor{agentbg}
\quad ACC
& \textbf{19.5${\pm}$1.1\%} & \textbf{0.096} & \$9.07 & 4.3\%
& \textbf{19.9${\pm}$1.9\%} & \textbf{0.106} & \$7.52 & 4.7\% \\
\bottomrule[1.2pt]
\end{tabularx}
\caption{
Main results for information fidelity under compression.
Flip and TVD are averaged across the two decision models; the ${\pm}$ on Flip is the s.d.\ across the two models.
\textit{No compression} is the identity floor, where the decision model rereads the source.
Flip denotes the rate at which compression changes the source-based decision.
TVD denotes source-relative probability drift.
Cost denotes total inference cost.
}
\label{tab:main-compression-results}
\end{table*}

\subsection{Decontextualization}
\label{subsec:fact-inventory}

We diagnose decontextualization with an offline fact inventory following FActScore and SAFE \citep{min2023factscore,wei2024safe}; details of the decomposition and role-labeling prompts are in Appendix~\ref{app:fact-decomposition}. We split each source disclosure into atomic facts, link each fact to its source sentence, and tag it as headline, context, or boilerplate. Headline facts carry the main investment signal; context facts calibrate that signal through caveats, offsets, comparisons, or expectation framing; boilerplate facts are background or procedural text.

\begin{figure}[htbp]
    \centering
    \includegraphics[width=0.54\linewidth]{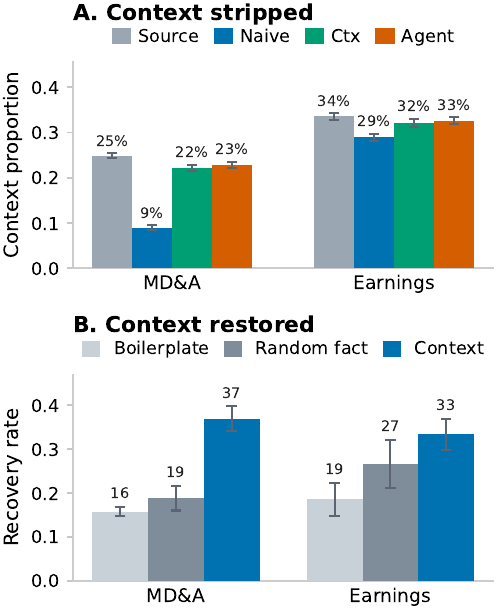}
    \caption{\textbf{(A)} Context facts are dropped much more under MD\&A than earnings-call. \textbf{(B)} Re-adding dropped context recovers more flipped decisions than boilerplate or random facts.}
    \label{fig:decontext}
\end{figure}

The second diagnostic pattern is decontextualization: under a fixed budget, compression often keeps headline results while dropping the caveats, offsets, comparisons, and forward-looking qualifiers needed to interpret them. Figure~\ref{fig:decontext}A shows that this stripping is selective: under MD\&A compression, the context share falls from 25\% in the source to 9\% in the one-shot compression. This disproportionate loss can make a factually accurate summary read less calibrated than the source it came from. The pattern is milder for shorter earnings transcripts, where more facts fit under the same budget and context is largely preserved.

We then run an add-back diagnostic on flip cases (Figure~\ref{fig:decontext}B). Starting from one-shot summaries that flipped the decision model's output, we append omitted facts and have the model reread the summary. Restoring dropped context facts recovers the source decision in 37\% of MD\&A flips, compared with 16--19\% for boilerplate or random facts from the same filing. Both placebo conditions append the same number of facts per document as the context condition, so the recovery gap reflects which facts are restored rather than how much text is added. Earnings calls show the same ordering but a narrower gap: 33\% for context versus 19--27\% for the placebos. This pattern is consistent with missing context being an important contributor, though the add-back exceeds the budget and remains diagnostic rather than deployable.

The gain is largest where the decontextualization diagnostic is strongest: for 10-Q MD\&A, where dropped context produces the clearest add-back recovery, contextualization lowers the flip rate from 31.5\% to 21.2\%. For earnings calls, where context is already better preserved under the budget, the gain is smaller, from 24.9\% to 22.1\%.

\subsection{Source-Grounded Selection and Cost}
\label{subsec:acc-selection}

\begin{table}[t]
\centering
\footnotesize
\renewcommand{\arraystretch}{1.12}
\resizebox{\columnwidth}{!}{%
\begin{tabular}{@{}lcc@{}}
\toprule[1.2pt]
Selection variant & 10-Q MD\&A & Earnings Call \\
\midrule
Contextualization A (\modelname{GPT}) & 21.3 [16.7, 26.0] & 21.5 [17.2, 26.3] \\
Contextualization B (\modelname{DeepSeek}) & 20.3 [15.7, 25.0] & 22.6 [17.8, 27.3] \\
Random candidate & 22.3 [17.7, 27.3] & 22.2 [17.8, 26.9] \\
ACC (audited selection) & \textbf{20.3} [15.7, 25.0] & \textbf{18.5} [14.1, 22.9] \\
\bottomrule[1.2pt]
\end{tabular}}
\caption{Ablation of the selection rule. The two candidate compressions are held fixed, and each row changes only how the final compressed context is chosen. Cells report Flip (\%) with 95\% bootstrap confidence intervals over documents.}
\label{tab:acc-ablation}
\end{table}

Table~\ref{tab:acc-ablation} holds ACC's two candidate compressions fixed and varies only the selection rule. Audited selection lowers the flip rate over a seeded random choice by 2.0pp on MD\&A and 3.7pp on earnings calls, and on MD\&A it matches the best fixed candidate. ACC recovers the best candidate without knowing it in advance (worked example in Appendix~\ref{app:worked-example}). Paired comparisons (Appendix~\ref{app:bootstrap}) show that the ACC$-$Naive reduction is significant on both source types and the audited-versus-random gain on earnings calls, while the ACC$-$Contextualization difference is not.

ACC carries cost overhead relative to a single compression pass. Against the budget-matched Integrator that consumes the same two compressors, it achieves lower flip rates at comparable aggregate cost. Because compressed representations are commonly reused downstream, fidelity errors can propagate beyond the initial call, so the additional audit cost is small relative to the cost of decisions made on distorted context.

\section{Industry Case Study}

To test whether these fidelity effects translate to production forecasting systems, we apply ACC in a globally deployed commercial equity-forecasting product used by institutional investors. The study is conducted on S\&P 100 disclosures from fiscal 2025 Q1--Q3. Forecasting quality is measured by the information coefficient (IC), the cross-sectional Spearman correlation between predicted and realized next-quarter returns, holding the forecasting model, stock universe, and evaluation period fixed and varying only the disclosure representation.

Relative to the original disclosures, both naive compression and ACC reduced token consumption by more than 90\%. However, naive compression degraded forecasting IC, suggesting that compression-induced decision changes remove forecast-relevant signals. In contrast, ACC improved forecasting IC by 8.3\% over the original-source baseline and by 23.8\% relative to naive compression, while reducing source-relative decision flips by 59.5\% (a relative reduction versus naive compression, not the rate of distorted outputs in production). In production, outputs that change the source-based decision are routed through ACC, and the original disclosure is retained when the discrepancy remains unresolved.

Preserving decision-relevant context thus matters not only for compression quality, but also for retaining downstream forecasting signal.

\section{Conclusion}
This paper examines fidelity loss in LLM context compression for financial decision-making: compressed contexts can remain fluent and factually plausible while still changing downstream investment judgments. Our results point to two recurring sources of such loss. One is decontextualization: when key caveats, offsets, or comparative framing are omitted, preserving isolated facts is insufficient. The other is model dependency: compressors shift decisions on the same source in different directions and magnitudes, so a single-model summary can carry a model-specific tilt. To address both, we propose ACC, which audits contextualized candidates from different models against the source and adopts the best-supported one. Under the same budget, ACC better preserves source-relative decisions than one-shot baselines. Overall, compression should be evaluated not only by efficiency or factuality, but also by its ability to preserve decision-relevant context.

\section{Limitations}
Our experiments focus on financial disclosures from S\&P 100 firms, so the findings may not directly generalize to other domains, firm types, or market conditions. In addition, fidelity is measured through decision models rather than human analysts or realized market outcomes, although the main conclusions are consistent across two independent decision models and a refined label space. Future work should validate these findings on broader domains and against human judgments.

\section*{Acknowledgments}
This work was supported by the National Research Foundation of Korea (NRF) grant funded by the Korea government (MSIT) (No. RS-2025-24803208) and the Institute of Information \& Communications Technology Planning \& Evaluation (IITP) grant funded by the Korea government (MSIT) (No. RS-2020-II201336, Artificial Intelligence Graduate School Program (UNIST)).

\bibliography{reference}

@inproceedings{mohamed2025broken,
  title = {{LLM} as a Broken Telephone: Iterative Generation Distorts Information},
  author = {Mohamed, Amr and Geng, Mingmeng and Vazirgiannis, Michalis and Shang, Guokan},
  editor = {Che, Wanxiang and Nabende, Joyce and Shutova, Ekaterina and Pilehvar, Mohammad Taher},
  booktitle = {Proceedings of the 63rd Annual Meeting of the Association for Computational Linguistics (Volume 1: Long Papers)},
  month = jul,
  year = {2025},
  address = {Vienna, Austria},
  publisher = {Association for Computational Linguistics},
  url = {https://aclanthology.org/2025.acl-long.371/},
  doi = {10.18653/v1/2025.acl-long.371},
  pages = {7493--7509},
  ISBN = {979-8-89176-251-0}
}

@inproceedings{zhang2026ace,
  title = {Agentic Context Engineering: Evolving Contexts for Self-Improving Language Models},
  author = {Zhang, Qizheng and Hu, Changran and Upasani, Shubhangi and Ma, Boyuan and Hong, Fenglu and Kamanuru, Vamsidhar and Rainton, Jay and Wu, Chen and Ji, Mengmeng and Li, Hanchen and Thakker, Urmish and Zou, James and Olukotun, Kunle},
  booktitle = {The Fourteenth International Conference on Learning Representations},
  year = {2026},
  url = {https://openreview.net/forum?id=eC4ygDs02R}
}

@misc{laban2026corrupt,
  title = {{LLMs Corrupt Your Documents When You Delegate}},
  author = {Laban, Philippe and Schnabel, Tobias and Neville, Jennifer},
  year = {2026},
  eprint = {2604.15597},
  archivePrefix = {arXiv}
}

@inproceedings{min2023factscore,
  title = {{FActScore: Fine-grained Atomic Evaluation of Factual Precision in Long Form Text Generation}},
  author = {Min, Sewon and Krishna, Kalpesh and Lyu, Xinxi and Lewis, Mike and Yih, Wen-tau and Koh, Pang and Iyyer, Mohit and Zettlemoyer, Luke and Hajishirzi, Hannaneh},
  editor = {Bouamor, Houda and Pino, Juan and Bali, Kalika},
  booktitle = {Proceedings of the 2023 Conference on Empirical Methods in Natural Language Processing},
  month = dec,
  year = {2023},
  address = {Singapore},
  publisher = {Association for Computational Linguistics},
  url = {https://aclanthology.org/2023.emnlp-main.741/},
  doi = {10.18653/v1/2023.emnlp-main.741},
  pages = {12076--12100}
}

@inproceedings{wei2024safe,
  title = {{Long-form Factuality in Large Language Models}},
  author = {Wei, Jerry and Yang, Chengrun and Song, Xinying and Lu, Yifeng and Hu, Nathan and Huang, Jie and Tran, Dustin and Peng, Daiyi and Liu, Ruibo and Huang, Da and Du, Cosmo and Le, Quoc V.},
  booktitle = {Advances in Neural Information Processing Systems},
  volume = {37},
  pages = {80756--80827},
  publisher = {Curran Associates, Inc.},
  year = {2024},
  url = {https://proceedings.neurips.cc/paper_files/paper/2024/hash/937ae0e83eb08d2cb8627fe1def8c751-Abstract-Conference.html}
}

@inproceedings{zhang2024fairsumm,
  title = {{Fair Abstractive Summarization of Diverse Perspectives}},
  author = {Zhang, Yusen and Zhang, Nan and Liu, Yixin and Fabbri, Alexander and Liu, Junru and Kamoi, Ryo and Lu, Xiaoxin and Xiong, Caiming and Zhao, Jieyu and Radev, Dragomir and McKeown, Kathleen and Zhang, Rui},
  editor = {Duh, Kevin and Gomez, Helena and Bethard, Steven},
  booktitle = {Proceedings of the 2024 Conference of the North American Chapter of the Association for Computational Linguistics: Human Language Technologies (Volume 1: Long Papers)},
  month = jun,
  year = {2024},
  address = {Mexico City, Mexico},
  publisher = {Association for Computational Linguistics},
  url = {https://aclanthology.org/2024.naacl-long.187/},
  doi = {10.18653/v1/2024.naacl-long.187},
  pages = {3404--3426}
}

@inproceedings{lei2024poca,
  title = {{Polarity Calibration for Opinion Summarization}},
  author = {Lei, Yuanyuan and Song, Kaiqiang and Cho, Sangwoo and Wang, Xiaoyang and Huang, Ruihong and Yu, Dong},
  editor = {Duh, Kevin and Gomez, Helena and Bethard, Steven},
  booktitle = {Proceedings of the 2024 Conference of the North American Chapter of the Association for Computational Linguistics: Human Language Technologies (Volume 1: Long Papers)},
  month = jun,
  year = {2024},
  address = {Mexico City, Mexico},
  publisher = {Association for Computational Linguistics},
  url = {https://aclanthology.org/2024.naacl-long.291/},
  doi = {10.18653/v1/2024.naacl-long.291},
  pages = {5211--5224}
}

@misc{aghaebe2026faithful,
  title = {{Faithful Summarisation under Disagreement via Belief-Level Aggregation}},
  author = {Aghaebe, Favour Yahdii and Apekey, Tanefa and Williams, Elizabeth and Moosavi, Nafise Sadat},
  year = {2026},
  eprint = {2601.04889},
  archivePrefix = {arXiv}
}

@misc{glasserman2023lookahead,
  title = {{Assessing Look-Ahead Bias in Stock Return Predictions Generated By GPT Sentiment Analysis}},
  author = {Glasserman, Paul and Lin, Caden},
  year = {2023},
  eprint = {2309.17322},
  archivePrefix = {arXiv}
}

@inproceedings{nakagawa2024company,
  title = {{Evaluating Company-specific Biases in Financial Sentiment Analysis using Large Language Models}},
  author = {Nakagawa, Kei and Hirano, Masanori and Fujimoto, Yugo},
  booktitle = {2024 IEEE International Conference on Big Data (BigData)},
  pages = {6614--6623},
  publisher = {IEEE},
  doi = {10.1109/BigData62323.2024.10826008},
  url = {https://doi.org/10.1109/BigData62323.2024.10826008},
  year = {2024},
  archivePrefix = {arXiv},
  arxivId = {2411.00420}
}

@inproceedings{lee2025yourai,
  title = {{Your AI, Not Your View: The Bias of LLMs in Investment Analysis}},
  author = {Lee, Hoyoung and Seo, Junhyuk and Park, Suhwan and Lee, Junhyeong and Ahn, Wonbin and Choi, Chanyeol and Lopez-Lira, Alejandro and Lee, Yongjae},
  booktitle = {Proceedings of the 6th ACM International Conference on AI in Finance},
  series = {ICAIF '25},
  pages = {150--158},
  location = {Singapore, Singapore},
  publisher = {Association for Computing Machinery},
  address = {New York, NY, USA},
  doi = {10.1145/3768292.3770375},
  url = {https://doi.org/10.1145/3768292.3770375},
  year = {2025},
  isbn = {979-8-4007-2220-2}
}

@inproceedings{kong2026bias,
  title = {{Evaluating LLMs in Finance Requires Explicit Bias Consideration}},
  author = {Kong, Yaxuan and Lee, Hoyoung and Hwang, Yoontae and Lopez-Lira, Alejandro and Levy, Bradford and Mehta, Dhagash and Wen, Qingsong and Choi, Chanyeol and Lee, Yongjae and Zohren, Stefan},
  booktitle = {Forty-third International Conference on Machine Learning},
  year = {2026},
  url = {https://openreview.net/forum?id=EDsAEXBFBk}
}

@inproceedings{loukas2025edgarcrawler,
  title = {{{EDGAR}-{CRAWLER}: From Raw Web Documents to Structured Financial NLP Datasets}},
  author = {Loukas, Lefteris and Billert, Fabian and Fergadiotis, Manos and Malakasiotis, Prodromos and Androutsopoulos, Ion},
  booktitle = {Companion Proceedings of the ACM on Web Conference 2025},
  series = {WWW '25},
  pages = {761--764},
  numpages = {4},
  location = {Sydney NSW, Australia},
  publisher = {Association for Computing Machinery},
  address = {New York, NY, USA},
  doi = {10.1145/3701716.3715289},
  url = {https://doi.org/10.1145/3701716.3715289},
  isbn = {979-8-4007-1331-6},
  year = {2025}
}

@misc{li2026dci,
  title = {Beyond Semantic Similarity: Rethinking Retrieval for Agentic Search via Direct Corpus Interaction},
  author = {Li, Zhuofeng and Zhang, Haoxiang and Wei, Cong and Lu, Pan and Nie, Ping and Lu, Yi and Bai, Yuyang and Feng, Shangbin and Zhu, Hangxiao and Zhong, Ming and Zhang, Yuyu and Xie, Jianwen and Choi, Yejin and Zou, James and Han, Jiawei and Chen, Wenhu and Lin, Jimmy and Jiang, Dongfu and Zhang, Yu},
  year = {2026},
  eprint = {2605.05242},
  archivePrefix = {arXiv}
}

@inproceedings{jiang2023llmlingua,
  title = {{LLMLingua}: Compressing Prompts for Accelerated Inference of Large Language Models},
  author = {Jiang, Huiqiang and Wu, Qianhui and Lin, Chin-Yew and Yang, Yuqing and Qiu, Lili},
  editor = {Bouamor, Houda and Pino, Juan and Bali, Kalika},
  booktitle = {Proceedings of the 2023 Conference on Empirical Methods in Natural Language Processing},
  month = dec,
  year = {2023},
  address = {Singapore},
  publisher = {Association for Computational Linguistics},
  url = {https://aclanthology.org/2023.emnlp-main.825/},
  doi = {10.18653/v1/2023.emnlp-main.825},
  pages = {13358--13376}
}

@inproceedings{jiang2024longllmlingua,
  title = {{LongLLMLingua}: Accelerating and Enhancing {LLMs} in Long Context Scenarios via Prompt Compression},
  author = {Jiang, Huiqiang and Wu, Qianhui and Luo, Xufang and Li, Dongsheng and Lin, Chin-Yew and Yang, Yuqing and Qiu, Lili},
  editor = {Ku, Lun-Wei and Martins, Andre and Srikumar, Vivek},
  booktitle = {Proceedings of the 62nd Annual Meeting of the Association for Computational Linguistics (Volume 1: Long Papers)},
  month = aug,
  year = {2024},
  address = {Bangkok, Thailand},
  publisher = {Association for Computational Linguistics},
  url = {https://aclanthology.org/2024.acl-long.91/},
  doi = {10.18653/v1/2024.acl-long.91},
  pages = {1658--1677}
}

@inproceedings{lajewska2025preservation,
  title = {{Understanding and Improving Information Preservation in Prompt Compression for LLMs}},
  author = {{\L}ajewska, Weronika and Hardalov, Momchil and Aina, Laura and Anna John, Neha and Su, Hang and Marquez, Lluis},
  editor = {Christodoulopoulos, Christos and Chakraborty, Tanmoy and Rose, Carolyn and Peng, Violet},
  booktitle = {Findings of the Association for Computational Linguistics: EMNLP 2025},
  month = nov,
  year = {2025},
  address = {Suzhou, China},
  publisher = {Association for Computational Linguistics},
  url = {https://aclanthology.org/2025.findings-emnlp.949/},
  doi = {10.18653/v1/2025.findings-emnlp.949},
  pages = {17520--17541},
  ISBN = {979-8-89176-335-7}
}

@inproceedings{mayilvaghanan2025blindspot,
  title = {{Spot the BlindSpots: Systematic Identification and Quantification of Fine-Grained LLM Biases in Contact Center Call Summarization}},
  author = {Mayilvaghanan, Kawin and Gupta, Siddhant and Kumar, Ayush},
  editor = {Potdar, Saloni and Rojas-Barahona, Lina and Montella, Sebastien},
  booktitle = {Proceedings of the 2025 Conference on Empirical Methods in Natural Language Processing: Industry Track},
  month = nov,
  year = {2025},
  address = {Suzhou (China)},
  publisher = {Association for Computational Linguistics},
  url = {https://aclanthology.org/2025.emnlp-industry.91/},
  doi = {10.18653/v1/2025.emnlp-industry.91},
  pages = {1299--1340},
  ISBN = {979-8-89176-333-3}
}

@inproceedings{kang2025acon,
  title = {{ACON: Optimizing Context Compression for Long-horizon LLM Agents}},
  author = {Kang, Minki and Chen, Wei-Ning and Han, Dongge and Inan, Huseyin A. and Wutschitz, Lukas and Chen, Yanzhi and Sim, Robert and Rajmohan, Saravan},
  booktitle = {Forty-third International Conference on Machine Learning},
  year = {2026},
  url = {https://openreview.net/forum?id=5EmOOLtH5P}
}

@article{choi2021decontext,
  title = {{Decontextualization: Making Sentences Stand-Alone}},
  author = {Choi, Eunsol and Palomaki, Jennimaria and Lamm, Matthew and Kwiatkowski, Tom and Das, Dipanjan and Collins, Michael},
  journal = {Transactions of the Association for Computational Linguistics},
  volume = {9},
  pages = {447--461},
  year = {2021},
  publisher = {MIT Press},
  doi = {10.1162/tacl_a_00377}
}

@inproceedings{gunjal2024molecular,
  title = {{Molecular Facts: Desiderata for Decontextualization in LLM Fact Verification}},
  author = {Gunjal, Anisha and Durrett, Greg},
  editor = {Al-Onaizan, Yaser and Bansal, Mohit and Chen, Yun-Nung},
  booktitle = {Findings of the Association for Computational Linguistics: EMNLP 2024},
  month = nov,
  year = {2024},
  address = {Miami, Florida, USA},
  publisher = {Association for Computational Linguistics},
  url = {https://aclanthology.org/2024.findings-emnlp.215/},
  doi = {10.18653/v1/2024.findings-emnlp.215},
  pages = {3751--3768}
}

@inproceedings{kuwahara2025conformal,
  title = {{Document Summarization with Conformal Importance Guarantees}},
  author = {Kuwahara, Bruce and Lin, Chen-Yuan and Huang, Xiao Shi and Leung, Kin Kwan and Yapeter, Jullian Arta and Stanevich, Ilya and P{\'e}rez, Felipe and Cresswell, Jesse C.},
  booktitle = {Advances in Neural Information Processing Systems},
  volume = {38},
  pages = {67107--67152},
  publisher = {Curran Associates, Inc.},
  year = {2025},
  doi = {10.52202/085713-2249},
  url = {https://proceedings.neurips.cc/paper_files/paper/2025/hash/60cefc59610f8f59ea10099f99a36726-Abstract-Conference.html}
}

@inproceedings{trienes2025salience,
  title = {{Behavioral Analysis of Information Salience in Large Language Models}},
  author = {Trienes, Jan and Schl{\"o}tterer, J{\"o}rg and Li, Junyi Jessy and Seifert, Christin},
  editor = {Che, Wanxiang and Nabende, Joyce and Shutova, Ekaterina and Pilehvar, Mohammad Taher},
  booktitle = {Findings of the Association for Computational Linguistics: ACL 2025},
  month = jul,
  year = {2025},
  address = {Vienna, Austria},
  publisher = {Association for Computational Linguistics},
  url = {https://aclanthology.org/2025.findings-acl.1204/},
  doi = {10.18653/v1/2025.findings-acl.1204},
  pages = {23428--23454},
  ISBN = {979-8-89176-256-5}
}

@misc{guo2026finground,
  title = {{FinGround: Detecting and Grounding Financial Hallucinations via Atomic Claim Verification}},
  author = {Guo, Dongxin and Wu, Jikun and Yiu, Siu Ming},
  year = {2026},
  eprint = {2604.23588},
  archivePrefix = {arXiv}
}

@misc{ning2025madfact,
  title = {{MAD-Fact: A Multi-Agent Debate Framework for Long-Form Factuality Evaluation in LLMs}},
  author = {Ning, Yucheng and Lin, Xixun and Fang, Fang and Cao, Yanan},
  year = {2025},
  eprint = {2510.22967},
  archivePrefix = {arXiv}
}

@inproceedings{alessa2025cognitive,
  title = {{Quantifying Cognitive Bias Induction in LLM-Generated Content}},
  author = {Alessa, Abeer and Somane, Param and Lakshminarasimhan, Akshaya Thenkarai and Skirzynski, Julian and McAuley, Julian and Echterhoff, Jessica Maria},
  editor = {Inui, Kentaro and Sakti, Sakriani and Wang, Haofen and Wong, Derek F. and Bhattacharyya, Pushpak and Banerjee, Biplab and Ekbal, Asif and Chakraborty, Tanmoy and Singh, Dhirendra Pratap},
  booktitle = {Proceedings of the 14th International Joint Conference on Natural Language Processing and the 4th Conference of the Asia-Pacific Chapter of the Association for Computational Linguistics},
  month = dec,
  year = {2025},
  address = {Mumbai, India},
  publisher = {The Asian Federation of Natural Language Processing and The Association for Computational Linguistics},
  url = {https://aclanthology.org/2025.ijcnlp-long.155/},
  doi = {10.18653/v1/2025.ijcnlp-long.155},
  pages = {2890--2910},
  ISBN = {979-8-89176-298-5}
}

@article{peters2025generalization,
  title = {{Generalization Bias in Large Language Model Summarization of Scientific Research}},
  author = {Peters, Uwe and Chin-Yee, Benjamin},
  journal = {Royal Society Open Science},
  volume = {12},
  number = {4},
  pages = {241776},
  year = {2025},
  publisher = {The Royal Society},
  doi = {10.1098/rsos.241776}
}

@inproceedings{pastorino2025framing,
  title = {{Frame In, Frame Out: Measuring Framing Bias in LLM-Generated News Summaries}},
  author = {Pastorino, Valeria and Moosavi, Nafise Sadat},
  editor = {Mohammad, Saif M. and Ousidhoum, Nedjma},
  booktitle = {Proceedings of the 15th Joint Conference on Lexical and Computational Semantics (*SEM 2026)},
  month = jul,
  year = {2026},
  address = {San Diego, California, United States},
  publisher = {Association for Computational Linguistics},
  url = {https://aclanthology.org/2026.starsem-conference.25/},
  doi = {10.18653/v1/2026.starsem-conference.25},
  pages = {378--384},
  ISBN = {979-8-89176-413-2}
}

@misc{zhang2025scope,
  title = {{SCOPE: A Generative Approach for LLM Prompt Compression}},
  author = {Zhang, Tinghui and Wang, Yifan and Wang, Daisy Zhe},
  year = {2025},
  eprint = {2508.15813},
  archivePrefix = {arXiv}
}

@inproceedings{lee2025theme,
  title = {{THEME: Enhancing Thematic Investing with Semantic Stock Representations and Temporal Dynamics}},
  author = {Lee, Hoyoung and Ahn, Wonbin and Park, Suhwan and Lee, Jaehoon and Kim, Minjae and Yoo, Sungdong and Lim, Taeyoon and Lim, Woohyung and Lee, Yongjae},
  booktitle = {Proceedings of the 34th ACM International Conference on Information and Knowledge Management},
  series = {CIKM '25},
  pages = {5797--5804},
  location = {Seoul, Republic of Korea},
  publisher = {Association for Computing Machinery},
  address = {New York, NY, USA},
  doi = {10.1145/3746252.3761517},
  url = {https://doi.org/10.1145/3746252.3761517},
  year = {2025}
}

@inproceedings{lee2026fintexts,
  title = {{FinTexTS: Financial Text-Paired Time-Series Dataset via Semantic-Based and Multi-Level Pairing}},
  author = {Lee, Jaehoon and Park, Suhwan and Lim, Taeyoon and Lee, Seunghan and Seo, Jun and Kang, Dongwan and Choi, Hwanil and Kim, Minjae and Yoo, Sungdong and Lee, Soonyoung and Lee, Yongjae and Ahn, Wonbin},
  booktitle = {Proceedings of the 32nd ACM SIGKDD Conference on Knowledge Discovery and Data Mining V.2},
  series = {KDD '26},
  pages = {9266--9277},
  location = {Jeju Island, Republic of Korea},
  publisher = {Association for Computing Machinery},
  address = {New York, NY, USA},
  doi = {10.1145/3770855.3817468},
  url = {https://doi.org/10.1145/3770855.3817468},
  year = {2026}
}

@misc{kim2026leadlag,
  title = {{LLM as a Risk Manager: LLM Semantic Filtering for Lead-Lag Trading in Prediction Markets}},
  author = {Kim, Sumin and Kim, Minjae and Kwon, Jihoon and Kim, Yoon and Kagan, Nicole and Lee, Joo Won and Levy, Oscar and Lopez-Lira, Alejandro and Lee, Yongjae and Choi, Chanyeol},
  year = {2026},
  eprint = {2602.07048},
  archivePrefix = {arXiv}
}

@misc{choi2025alpha,
  title = {{From Text to Alpha: Can LLMs Track Evolving Signals in Corporate Disclosures?}},
  author = {Choi, Chanyeol and Kim, Yoon and Yu, Yu and Cha, Young and Golkhou, V. Zach and Halperin, Igor and Papaioannou, Georgios and Kim, Minkyu and Wang, Zhangyang and Kwon, Jihoon and Kim, Minjae and Lopez-Lira, Alejandro and Lee, Yongjae},
  year = {2025},
  eprint = {2510.03195},
  archivePrefix = {arXiv}
}

\appendix
\raggedbottom

\clearpage
\onecolumn
\section{Budget Sensitivity}
\label{app:budget-sensitivity}

As a brief robustness check, Figure~\ref{fig:budget-sensitivity} repeats one-shot compression with budgets from 5 to 40 bullets on a 150-document subset, evaluated under \modelname{Gemini-3.1-Flash-Lite} only. Larger budgets reduce source-relative TVD, especially from 20 to 40 bullets, but decision flips remain well above that model's no-compression floor. Extra space therefore smooths belief drift more reliably than it eliminates top-label changes.

\begin{center}
    \centering
\includegraphics[width=0.60\textwidth]{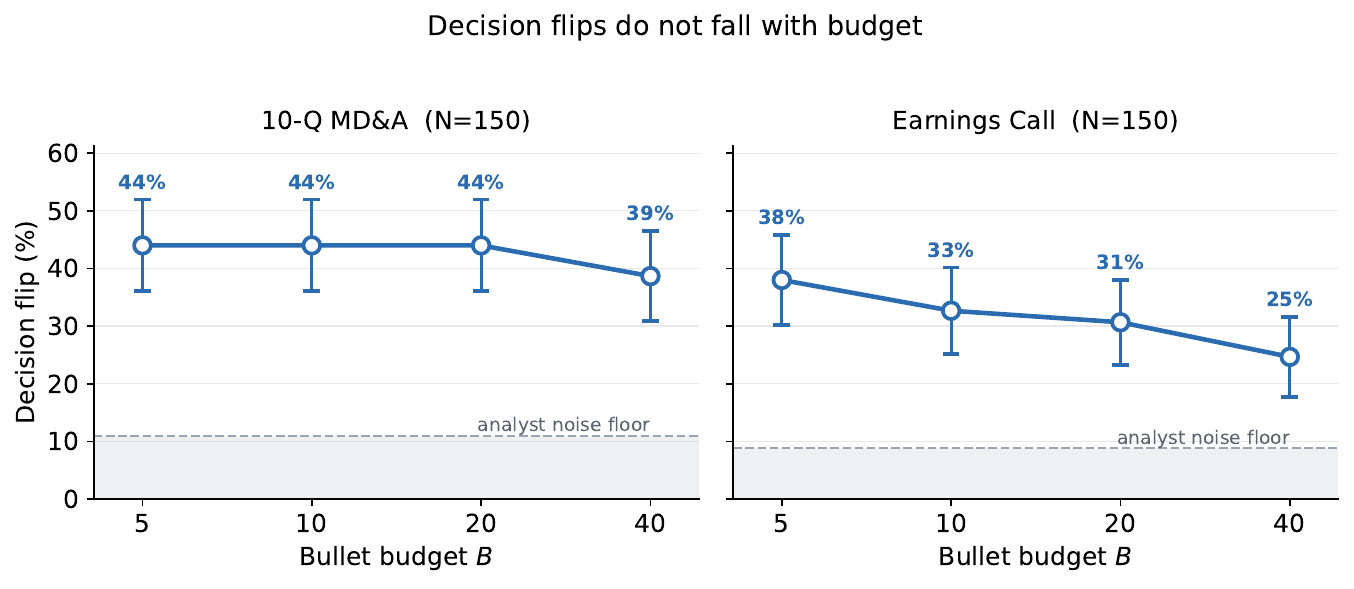}
    \vspace{0.15em}
\includegraphics[width=0.60\textwidth]{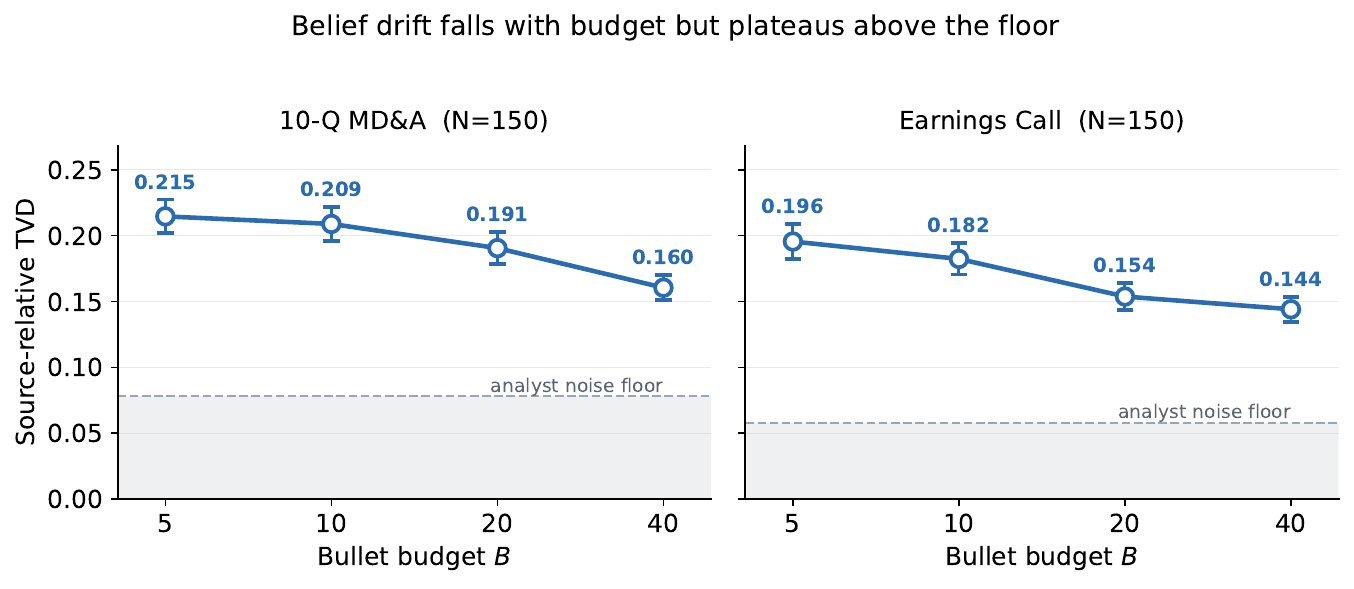}
    \captionof{figure}{Budget sensitivity under \modelname{GPT-5.4-Mini} one-shot compression. Top: decision flips remain above the decision-model noise floor even at larger budgets. Bottom: source-relative TVD declines with budget but plateaus above the floor.}
    \label{fig:budget-sensitivity}
\end{center}

\section{Inter-Model Agreement}
\label{app:model-agreement}

Aggregate flip rates can hide whether compressors fail on the same filings. Figure~\ref{fig:model-agreement-heatmap} reports pairwise top-decision agreement across four one-shot compressors on the 597 source documents pooled across MD\&A and earnings-call Q\&A. The mean off-diagonal agreement is 0.75, so compressors disagree on roughly one filing in four even when their marginal flip rates are similar. This supports treating compression fidelity as model-dependent rather than as a property of the source alone.

\begin{center}
    \centering
\includegraphics[width=0.46\textwidth]{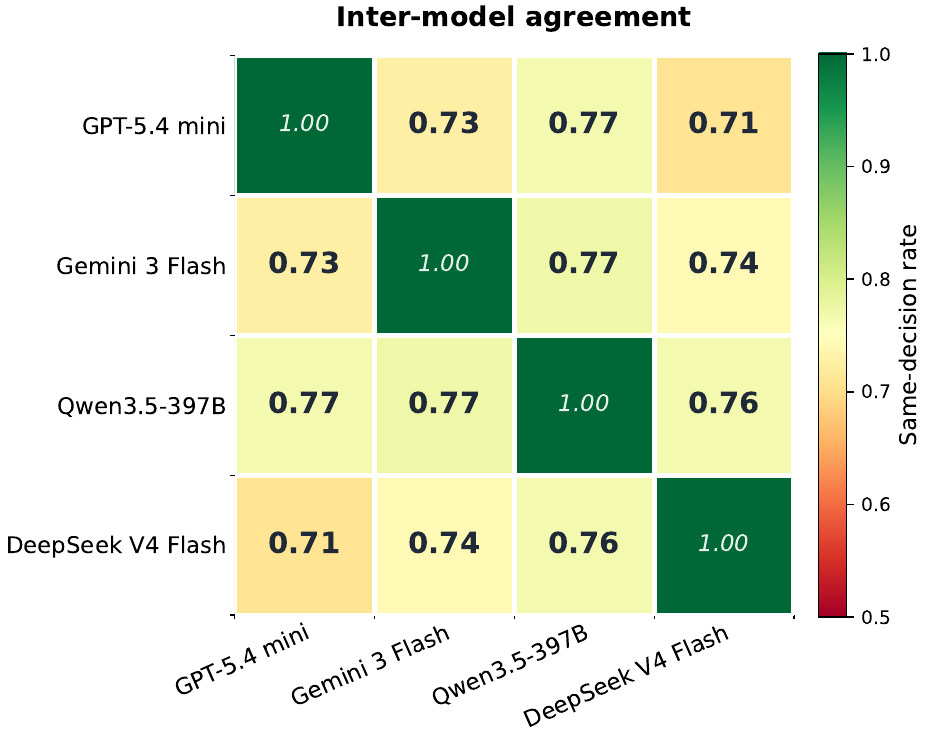}
    \captionof{figure}{Pairwise top-decision agreement among one-shot compressors on the same source filings. Cells show the share of filings for which two compressors induce the same downstream bear/neutral/bull decision; off-diagonal values near 0.75 indicate that compressor choice changes the induced decision on about one quarter of filings.}
    \label{fig:model-agreement-heatmap}
\end{center}

\clearpage
\section{Decision Movement}
\label{app:decision-movement-source-type}

\begin{figure}[!htbp]
    \centering
    \includegraphics[width=0.92\textwidth]{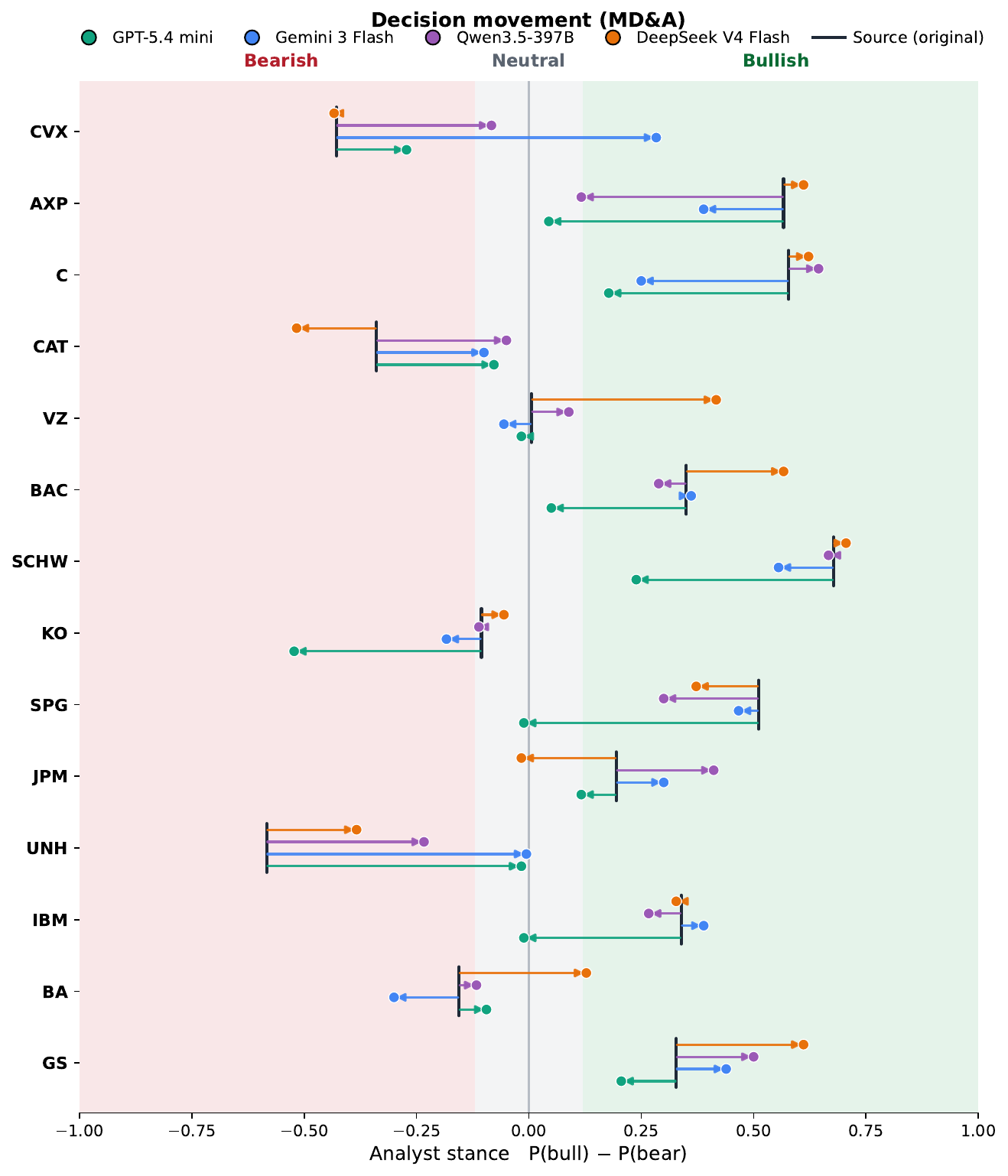}
    \caption{Decision movement for MD\&A disclosures.}
    \label{fig:decision-movement-mda}
\end{figure}

\begin{figure}[!htbp]
    \centering
    \includegraphics[width=0.92\textwidth]{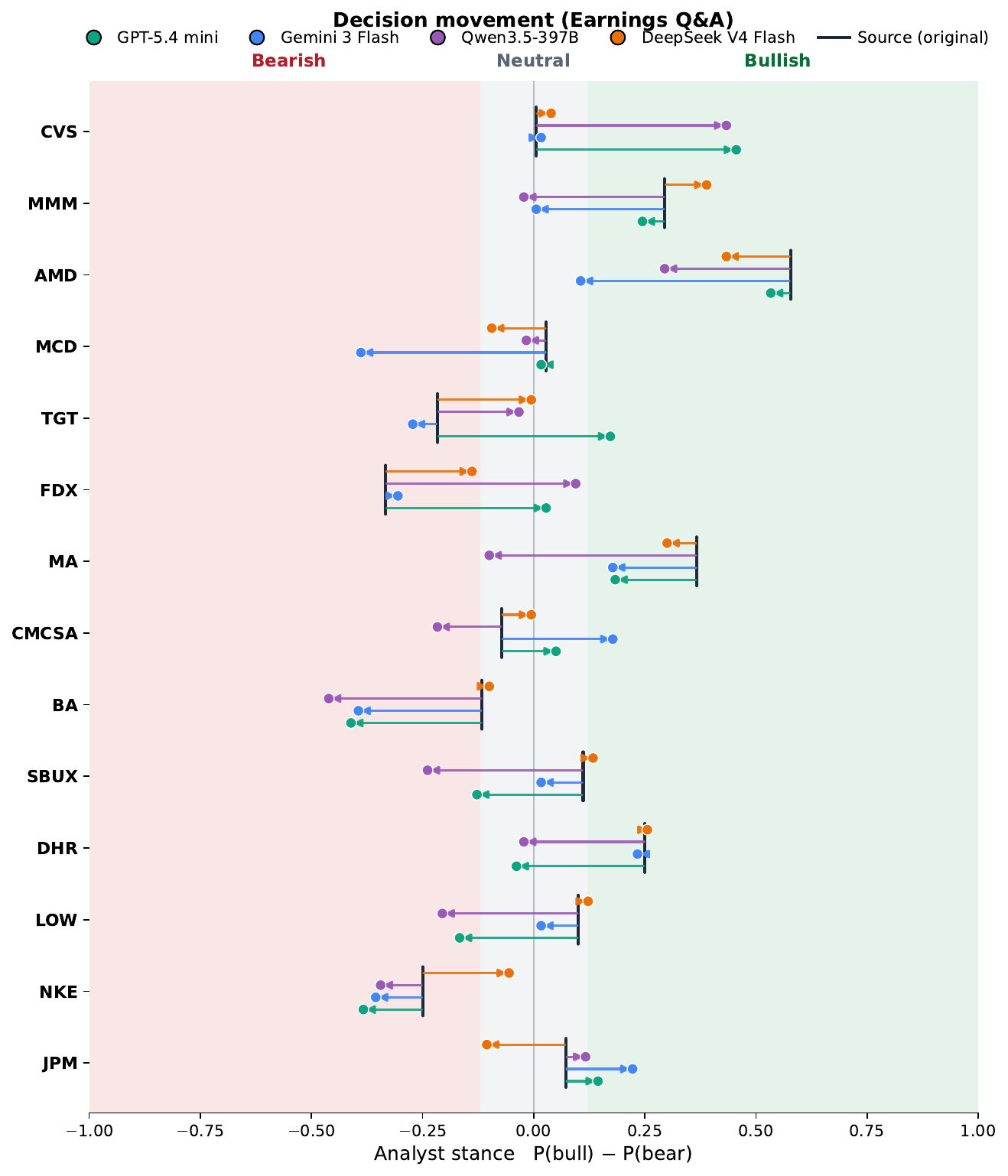}
    \caption{Decision movement for earnings-call Q\&A disclosures.}
    \label{fig:decision-movement-earnings-qa}
\end{figure}

Figures~\ref{fig:decision-movement-mda} and~\ref{fig:decision-movement-earnings-qa} show the same movement diagnostic separately for MD\&A and earnings-call Q\&A. The black vertical marker is the source-implied stance for each ticker, and each colored arrow shows where a one-shot compressor moves the downstream decision. Wider horizontal spread within a ticker indicates stronger model dependence.

\clearpage

\section{Per-Evaluator Results}
\label{app:per-evaluator}

Table~\ref{tab:per-evaluator} reports the components of Table~\ref{tab:main-compression-results} separately for the two decision models; Flip is shown with the s.d.\ across each model's three decision runs. The main contrasts hold under both evaluators: Contextualization and ACC lower the flip rate relative to the Naive prompt on both source types, and the token-pruning baselines remain weakest. The full ordering is not identical, however: on MD\&A, Chunking is the weakest LLM method under \modelname{Gemini-3.1-Flash-Lite} but outperforms the Naive prompt under \modelname{Claude-Haiku-4.5}, and on earnings calls the Integrator attains a lower flip rate than ACC under \modelname{Claude-Haiku-4.5} (20.5\% vs.\ 21.2\%), consistent with framing ACC as source-grounded selection. The candidate-A row of Table~\ref{tab:acc-ablation} re-evaluates the same artifacts under fresh decision reads; its gap from the 22.2\% entry here is decision-run resampling noise, well inside the bootstrap CI.

\begin{center}
\small
\renewcommand{\arraystretch}{1.12}
\begin{tabular}{lcccc}
\toprule[1.2pt]
& \multicolumn{2}{c}{10-Q MD\&A} & \multicolumn{2}{c}{Earnings Call} \\
\cmidrule(lr){2-3} \cmidrule(lr){4-5}
Method & Flip $\downarrow$ & TVD $\downarrow$ & Flip $\downarrow$ & TVD $\downarrow$ \\
\midrule
\rowcolor{groupbg}
\multicolumn{5}{l}{\modelname{Gemini-3.1-Flash-Lite}} \\
\midrule
\quad Naive Prompt & 33.0 ${\pm}$ 1.6 & 0.172 & 23.9 ${\pm}$ 1.4 & 0.137 \\
\quad Chunking & 37.0 ${\pm}$ 1.3 & 0.155 & 31.6 ${\pm}$ 0.5 & 0.147 \\
\quad Contextualization & 21.3 ${\pm}$ 1.8 & \textbf{0.099} & 22.2 ${\pm}$ 3.7 & 0.116 \\
\quad LLMLingua & 53.0 ${\pm}$ 0.7 & 0.231 & 50.8 ${\pm}$ 1.8 & 0.247 \\
\quad LongLLMLingua & 50.0 ${\pm}$ 1.0 & 0.226 & 39.1 ${\pm}$ 1.9 & 0.196 \\
\quad Integrator & 24.7 ${\pm}$ 1.0 & 0.128 & 22.9 ${\pm}$ 1.9 & 0.128 \\
\quad ACC & \textbf{20.3} ${\pm}$ 1.7 & 0.103 & \textbf{18.5} ${\pm}$ 0.4 & \textbf{0.111} \\
\midrule
\rowcolor{groupbg}
\multicolumn{5}{l}{\modelname{Claude-Haiku-4.5}} \\
\midrule
\quad Naive Prompt & 29.9 ${\pm}$ 0.9 & 0.135 & 25.9 ${\pm}$ 0.9 & 0.118 \\
\quad Chunking & 23.1 ${\pm}$ 1.2 & 0.111 & 27.9 ${\pm}$ 1.9 & 0.129 \\
\quad Contextualization & 21.1 ${\pm}$ 0.5 & 0.094 & 21.9 ${\pm}$ 1.2 & 0.106 \\
\quad LLMLingua & 46.6 ${\pm}$ 1.7 & 0.205 & 68.7 ${\pm}$ 3.7 & 0.306 \\
\quad LongLLMLingua & 36.1 ${\pm}$ 1.1 & 0.174 & 42.4 ${\pm}$ 1.2 & 0.186 \\
\quad Integrator & 26.5 ${\pm}$ 0.7 & 0.113 & \textbf{20.5} ${\pm}$ 1.6 & \textbf{0.100} \\
\quad ACC & \textbf{18.7} ${\pm}$ 2.1 & \textbf{0.089} & 21.2 ${\pm}$ 0.5 & 0.101 \\
\bottomrule[1.2pt]
\end{tabular}
\captionof{table}{Per-evaluator information fidelity results. Flip (\%) is reported ${\pm}$ the s.d.\ across the three decision runs of each decision model; TVD is the mean source-relative total variation distance.}
\label{tab:per-evaluator}
\end{center}

\section{Bootstrap Confidence Intervals and Paired Comparisons}
\label{app:bootstrap}

Table~\ref{tab:bootstrap-ci} reports 95\% bootstrap confidence intervals (5{,}000 document resamples) for the main methods under each decision model, and Table~\ref{tab:paired-deltas} reports paired flip deltas over the same documents. Under \modelname{Gemini-3.1-Flash-Lite}, the ACC$-$Naive reduction is significant on both source types and the audited-versus-random gain on earnings calls, while the ACC$-$Contextualization difference is not; \modelname{Claude-Haiku-4.5} shows the same qualitative pattern.

\begin{center}
\small
\renewcommand{\arraystretch}{1.12}
\setlength{\tabcolsep}{4.5pt}
\begin{tabular}{lcccc}
\toprule[1.2pt]
& \multicolumn{2}{c}{10-Q MD\&A} & \multicolumn{2}{c}{Earnings Call} \\
\cmidrule(lr){2-3} \cmidrule(lr){4-5}
Method & Flip [95\% CI] & TVD [95\% CI] & Flip [95\% CI] & TVD [95\% CI] \\
\midrule
\rowcolor{groupbg}
\multicolumn{5}{l}{\modelname{Gemini-3.1-Flash-Lite}} \\
\midrule
\quad Naive Prompt & 33.0 [27.7, 38.3] & 0.172 [0.157, 0.186] & 23.9 [19.2, 29.0] & 0.137 [0.124, 0.150] \\
\quad Contextualization & 21.3 [16.7, 26.0] & 0.098 [0.089, 0.108] & 22.2 [17.5, 27.3] & 0.116 [0.105, 0.127] \\
\quad Integrator & 24.7 [20.0, 29.7] & 0.128 [0.117, 0.140] & 22.9 [18.2, 27.6] & 0.128 [0.116, 0.140] \\
\quad ACC & 20.3 [15.7, 25.0] & 0.103 [0.093, 0.112] & 18.5 [14.1, 22.9] & 0.111 [0.101, 0.122] \\
\midrule
\rowcolor{groupbg}
\multicolumn{5}{l}{\modelname{Claude-Haiku-4.5}} \\
\midrule
\quad Naive Prompt & 29.9 [24.8, 35.0] & 0.135 [0.121, 0.148] & 25.9 [20.9, 31.0] & 0.118 [0.106, 0.131] \\
\quad Contextualization & 21.1 [16.7, 25.9] & 0.094 [0.082, 0.106] & 21.9 [17.2, 26.6] & 0.106 [0.095, 0.118] \\
\quad Integrator & 26.5 [21.8, 31.6] & 0.113 [0.100, 0.127] & 20.5 [15.8, 25.3] & 0.100 [0.089, 0.112] \\
\quad ACC & 18.7 [14.3, 23.1] & 0.089 [0.078, 0.100] & 21.2 [16.5, 25.9] & 0.101 [0.091, 0.113] \\
\bottomrule[1.2pt]
\end{tabular}
\captionof{table}{95\% bootstrap confidence intervals (5{,}000 document resamples) for Flip (\%) and TVD under each decision model.}
\label{tab:bootstrap-ci}
\end{center}

\begin{center}
\small
\renewcommand{\arraystretch}{1.12}
\begin{tabular}{lcc}
\toprule[1.2pt]
Contrast & 10-Q MD\&A & Earnings Call \\
\midrule
\rowcolor{groupbg}
\multicolumn{3}{l}{\modelname{Gemini-3.1-Flash-Lite}} \\
\midrule
\quad Naive $\rightarrow$ Contextualization & $-$11.7 [$-$17.7, $-$5.3] & $-$1.7 [$-$7.4, $+$4.0] \\
\quad Contextualization A $\rightarrow$ random candidate & $+$1.0 [$-$2.7, $+$4.7] & $+$0.7 [$-$3.4, $+$4.7] \\
\quad Random candidate $\rightarrow$ ACC & $-$2.0 [$-$5.3, $+$1.3] & $-$3.7 [$-$7.4, $-$0.3] \\
\quad Naive $\rightarrow$ ACC & $-$12.7 [$-$19.0, $-$6.3] & $-$5.4 [$-$10.4, $-$0.3] \\
\quad Contextualization $\rightarrow$ ACC & $-$1.0 [$-$5.7, $+$3.7] & $-$3.7 [$-$8.8, $+$1.0] \\
\quad Integrator $\rightarrow$ ACC & $-$4.3 [$-$10.0, $+$1.3] & $-$4.4 [$-$9.8, $+$0.7] \\
\midrule
\rowcolor{groupbg}
\multicolumn{3}{l}{\modelname{Claude-Haiku-4.5}} \\
\midrule
\quad Naive $\rightarrow$ Contextualization & $-$8.8 [$-$14.6, $-$3.1] & $-$4.0 [$-$9.8, $+$1.7] \\
\quad Naive $\rightarrow$ ACC & $-$11.2 [$-$17.0, $-$5.1] & $-$4.7 [$-$10.4, $+$1.0] \\
\quad Contextualization $\rightarrow$ ACC & $-$2.4 [$-$7.1, $+$2.4] & $-$0.7 [$-$4.4, $+$3.0] \\
\quad Integrator $\rightarrow$ ACC & $-$7.8 [$-$13.3, $-$2.4] & $+$0.7 [$-$4.7, $+$6.4] \\
\bottomrule[1.2pt]
\end{tabular}
\captionof{table}{Paired flip deltas in percentage points with 95\% bootstrap CIs over documents. The candidate-routing contrasts (rows two and three) are defined by the fixed-candidate ablation and reported under \modelname{Gemini-3.1-Flash-Lite} only.}
\label{tab:paired-deltas}
\end{center}

\section{Five-Class Label Space}
\label{app:five-class}

Keeping the compressed artifacts fixed, we re-run \modelname{Gemini-3.1-Flash-Lite} with five next-quarter return bands (strong bear, bear, neutral, bull, strong bull), averaging two independent decision runs. The main conclusion persists (Table~\ref{tab:five-class}); on MD\&A, contextualization attains the lowest five-class flip rate (18.7\% vs.\ 21.3\% for ACC), consistent with the source-grounded-selection framing.

\begin{center}
\small
\renewcommand{\arraystretch}{1.12}
\begin{tabular}{lcccc}
\toprule[1.2pt]
& \multicolumn{2}{c}{10-Q MD\&A} & \multicolumn{2}{c}{Earnings Call} \\
\cmidrule(lr){2-3} \cmidrule(lr){4-5}
Method & Flip $\downarrow$ & TVD $\downarrow$ & Flip $\downarrow$ & TVD $\downarrow$ \\
\midrule
Naive Prompt & 36.0 & 0.185 & 30.7 & 0.165 \\
Contextualization & \textbf{18.7} & \textbf{0.125} & 30.7 & 0.151 \\
Integrator & 26.7 & 0.151 & 27.3 & 0.161 \\
ACC & 21.3 & \textbf{0.125} & \textbf{23.3} & \textbf{0.149} \\
\bottomrule[1.2pt]
\end{tabular}
\captionof{table}{Information fidelity under a five-class next-quarter return label space, with the compressed artifacts held fixed and \modelname{Gemini-3.1-Flash-Lite} averaged over two independent decision runs.}
\label{tab:five-class}
\end{center}

\section{Worked Example of the ACC Audit}
\label{app:worked-example}

In this logged case from a Lockheed Martin earnings call, management describes the Golden Dome opportunity as fluid and not yet quantifiable and warns that charges on legacy programs cannot guarantee zero future issues; the two candidate compressions treat this evidence differently:

\begin{promptbox}{Worked audit example (Lockheed Martin, earnings call)}
\textbf{Candidate A} (\modelname{GPT-5.4-Mini}) writes:\par
- Munitions and Golden Dome are the main new upside.\par
\smallskip
\textbf{Candidate B} (\modelname{DeepSeek-V4-Flash}) writes:\par
- Golden Dome unquantifiable until architecture and budget set; Lockheed well-positioned.\par
- Management took charges on legacy helicopter and missile programs to reduce future risk, but cannot guarantee zero issues.\par
\smallskip
\textbf{Audit.} Candidate A elevates the fluid, not-yet-quantifiable opportunity into the main new upside and omits the legacy-program warning; the audit rates A as higher overclaim risk against the retrieved source spans. Candidate B retains both qualifications and is selected verbatim.\par
\smallskip
\textbf{Outcome.} Without observing any evaluator outputs, ACC selects candidate B. Under \modelname{Gemini-3.1-Flash-Lite}, the source and candidate B induce neutral decisions, whereas candidate A induces a bull decision.
\end{promptbox}

\clearpage
\section{Prompt Templates}
\label{app:prompts}

The boxed templates below reproduce the prompts used in our experiments. Template variables are instantiated with the values used for the main results: the bullet budget is $B{=}20$ and word caps are stated per box. Compression prompts are sent as system messages; the user message supplies the source document and, for the integrator, the model views.

\subsection{One-Shot Compression Prompt}

\begin{promptbox}{One-shot compression}
\textbf{System.}\par
Compress the document below into exactly 20 bullets. This count is STRICT; do not exceed it.\par
Each bullet must be on its own line starting with `- '. Do not write headers or blank lines.\par
Each bullet captures exactly ONE fact or figure. Do not combine multiple facts into a single bullet.\par
If the document contains more facts than 20, select only the 20 most material facts and omit the rest.\par
Output only the bullets.\par
\smallskip
\textbf{User.}\par
{}[\{ticker\}] \{company\} \{document\_label\}\par
<document>\par
\{source\}\par
</document>
\end{promptbox}

\subsection{Contextualization Compression Prompt}
\label{app:context-prompt}

\begin{promptbox}{Contextualization compression}
Compress the document below into exactly 20 bullets. This count is STRICT; do not exceed it.\par
Each bullet must be on its own line starting with `- '. Do not write headers or blank lines.\par
Each bullet captures exactly ONE decision-relevant point. Do not combine unrelated points into one bullet.\par
The 20 bullets together should be about 240-250 words total. Do not exceed 250 words, but do not make the summary much shorter unless the source lacks material decision-relevant information.\par
\smallskip
A downstream decision model will make a bear / neutral / bull investment judgment from your bullets alone. Your goal is to preserve the source's decision direction and confidence as closely as possible. The compressed bullets should induce the same overall investment interpretation as the full source.\par
\smallskip
Selection rules:\par
- Preserve the source's dominant signal when the source is directionally positive, negative, or neutral.\par
- Preserve facts that explain why that signal is strong, weak, improving, deteriorating, uncertain, or mixed.\par
- When including a number, keep the comparison or frame that gives it meaning: prior year, prior quarter, guidance, consensus, management expectation, segment trend, margin impact, liquidity impact, or risk exposure.\par
- Do not write standalone figures without explaining what they imply.\par
- Preserve management's framing when it changes interpretation.\par
- Preserve forward-looking guidance, demand signals, margin or cost pressure, capital allocation, liquidity, and material risks when they affect the investment judgment.\par
- Include caveats or offsetting evidence only when they materially change the interpretation or confidence.\par
- Do not mechanically include both positive and negative facts.\par
- Do not force neutrality if the source has a clear directional signal.\par
- Avoid boilerplate, immaterial details, and facts that are vivid but unlikely to affect the investment judgment.\par
\smallskip
Use only what the source actually says; do not add outside information.\par
Output only the bullets.
\end{promptbox}

\newpage

\subsection{Multi-LLM Integrator Prompt}

\begin{promptbox}{Multi-LLM integrator}
You are a view-only financial-disclosure compression integrator.\par
\smallskip
You will receive one-shot compressed summaries from different LLMs. Treat the summaries as compressed model views, not as truth. Integrate them into one 20-bullet summary that keeps the decision-relevant information visible across the views.\par
\smallskip
Rules:\par
- Use only the provided compressed model views and task metadata.\par
- Do not use source decision labels, decision-model probabilities, realized outcomes, or external knowledge.\par
- Do not take a majority vote across views.\par
- Do not invent facts to fill gaps that no view mentions.\par
- If views differ, use cautious wording or keep both material perspectives.\par
- Output exactly 20 bullets.\par
- Each bullet must start with `- ' and capture one view-supported fact or qualifier.\par
- Output only the bullets, with no heading or commentary.\par
- The entire 20-bullet summary must total at most 330 words. Keep every bullet concise; drop lower-value detail rather than exceed the word budget.
\end{promptbox}

\subsection{Agentic Context Compression}
\label{app:agent}

\begin{methodbox}{Agent runtime contract}
\textbf{Candidates.} The agent adjudicates two one-shot Contextualization summaries of the same disclosure. Candidate A and candidate B are produced independently by two compression models.\par
\smallskip
\textbf{Source access.} The raw \texttt{source.txt} is not placed in the prompt, and the agent does not read the disclosure end-to-end. It uses targeted tools: \texttt{get\_metadata} identifies the document and source length, \texttt{grep\_source} searches for terms or claims where candidates disagree, and \texttt{read\_source\_span} opens short local passages around retrieved evidence. These inspections are span-limited; the agent must perform at least four targeted source checks before deciding.\par
\smallskip
\textbf{Audit criteria.} The audit focuses on disagreements between candidates and checks them against retrieved source spans for \emph{overclaim risk} and \emph{source-locus coverage}. Each checked claim is labeled \emph{supported}, \emph{unclear}, or \emph{overclaiming} and each material source locus \emph{covered}, \emph{partial}, or \emph{missed}; these labels are aggregated into categorical overclaim-risk and coverage scores per candidate.\par
\smallskip
\textbf{Selection rule.} The agent must select exactly one candidate, A or B. It compares the two candidates using retrieved source spans, chooses the candidate with lower overclaim risk, and uses broader source-locus coverage as the deciding criterion when overclaim risk is comparable.
\end{methodbox}

\newpage
\subsection{Decision Model Prompt}

\begin{promptbox}{Decision model}
\textbf{System.}\par
You are a financial decision model. You are given a corporate disclosure (or a compressed summary of one). Do not use prior knowledge of the company or any external context.\par
\smallskip
Estimate the disclosure-implied update, not a full investment recommendation. Use ONLY the provided disclosure evidence. Do not use valuation, stock price, prior returns, external company knowledge, or market context.\par
\smallskip
Based on the disclosure evidence alone, estimate how the disclosure would update an investor's belief about the stock's NEXT-QUARTER total-return regime:\par
bear = disclosure evidence points to downside risk (< -5\% if realized)\par
neutral = evidence is mixed, immaterial, or balanced (-5\% to +5\% if realized)\par
bull = disclosure evidence points to upside potential (> +5\% if realized)\par
\smallskip
Output ONLY a JSON object with three probabilities that sum to 1.0, no other text:\par
\{"bear": 0.xx, "neutral": 0.xx, "bull": 0.xx\}\par
\smallskip
\textbf{User.}\par
Ticker: \{ticker\} (\{company\}, \{sector\})\par
Period of report: \{period\} (\{q\_label\} \{year\})\par
Disclosure type: \{level\_label\}\par
\smallskip
Disclosure:\par
\{material\}\par
\smallskip
Disclosure-implied probability distribution over next-quarter regime (JSON only):
\end{promptbox}

\subsection{Fact Decomposition and Role Labeling}
\label{app:fact-decomposition}

\begin{promptbox}{Fact decomposition}
You are a financial-document fact extractor. Decompose the 10-Q MD\&A below into atomic facts. For each fact, return both the atomic claim and the source sentence it was extracted from.\par
\smallskip
Rules:\par
- Each fact is ONE atomic claim or figure. Do not combine multiple claims.\par
- Preserve quantitative detail verbatim (numbers, percentages, dates, entities).\par
- "source" is the full sentence (or short adjacent span) from the MD\&A that contains the fact. Copy it VERBATIM from the document -- do not paraphrase or summarize.\par
- Do NOT assign sentiment, polarity, or positivity/negativity. Sentiment is labeled in a separate step.\par
- Skip boilerplate (forward-looking disclaimers, table-of-contents text, GAAP definitions) and strictly neutral structural sentences.\par
- Return ONLY a JSON object: \{"facts": [\{"content": str, "source": str\}]\}. No prose, no markdown fences.
\end{promptbox}

\newpage
\begin{promptbox}{Role labeling}
You are a financial fact-role labeling assistant.\par
\smallskip
You are labeling atomic facts from a financial disclosure for an information-fidelity study.\par
\smallskip
The study asks whether compression preserves the context needed to interpret headline investment signals.\par
\smallskip
You will receive a numbered list of atomic facts from one disclosure. For each atomic fact, assign exactly one evidence-role label:\par
\smallskip
Use exactly one label:\par
- "headline": a primary investment signal that could directly move the decision model's assessment. This includes standalone facts about revenue, sales, demand, margins, profitability, costs, cash flow, guidance, growth, decline, market share, customer traction, operational performance, capital allocation, liquidity, or material risk when the fact itself is the main directional signal.\par
- "context": a fact that calibrates how a headline should be read. Label a fact as context only if it changes the interpretation of a headline signal by providing a caveat or limitation, offsetting evidence, year-over-year/\allowbreak sequential/\allowbreak segment/\allowbreak geography/\allowbreak customer/\allowbreak product comparison, magnitude/\allowbreak baseline/\allowbreak expectation-relative framing, causal driver, timing/\allowbreak durability/\allowbreak one-time nature, forward-looking qualifier, or risk condition/\allowbreak dependency.\par
- "boilerplate": generic background, corporate description, accounting/legal boilerplate, procedural text, vague management language, or scene-setting that does not materially affect how a headline investment signal should be interpreted.\par
\smallskip
Important: Do not label a fact as context merely because it is detailed, neutral, hedged, explanatory, legal/regulatory, or generally relevant to the business. A context fact must help interpret the direction, strength, scope, durability, cause, or expectation-relative meaning of a headline.\par
\smallskip
Decision rule: Ask: "If this fact were removed, would the decision model interpret the relevant headline signal differently?" If yes, and the fact calibrates another signal, label it "context". If the fact itself is the main directional signal, label it "headline". If it does not affect interpretation of a decision-relevant signal, label it "boilerplate".\par
\smallskip
Tie-break rules:\par
1. If a fact contains both a headline result and a qualifier that changes how the result should be read, choose "context".\par
2. If it is a standalone KPI/\allowbreak result/\allowbreak metric/\allowbreak guidance/\allowbreak capital-allocation fact, choose "headline".\par
3. If it is generic, procedural, definitional, or only broadly relevant, choose "boilerplate".\par
\smallskip
Output exactly one JSON object and no prose:\par
\{"labels": [\{"fact\_id": 0, "role": "<headline|\allowbreak context|\allowbreak boilerplate>", "context\_function": "<none|\allowbreak caveat|\allowbreak offset|\allowbreak comparison|\allowbreak expectation\_framing|\allowbreak causal\_driver|\allowbreak timing\_durability|\allowbreak scope\_magnitude|\allowbreak risk\_condition>", "target\_headline": "<short description or null>", "rationale": "<one short sentence>"\}]\}
\end{promptbox}

\end{document}